\documentclass[11pt]{article} 
\usepackage{url}
\usepackage{smile}
\usepackage{CJKutf8}
\usepackage{graphicx} 
\usepackage{todonotes}
\usepackage{epstopdf}
\usepackage{wrapfig}
\usepackage[colorlinks, linkcolor=blue, anchorcolor=blue, citecolor=blue]{hyperref}
\usepackage[margin=1in]{geometry}
\usepackage[normalem]{ulem}
\usepackage[export]{adjustbox}
\usepackage{mathtools, cuted}
\usepackage{natbib}
\usepackage{enumerate}
\usepackage{microtype}
\usepackage{graphicx}
\usepackage{booktabs} 
\usepackage{wrapfig}
\usepackage{xspace}

\usepackage{kpfonts}
\DeclareMathAlphabet{\mathsf}{OT1}{cmss}{m}{n}
\SetMathAlphabet{\mathsf}{bold}{OT1}{cmss}{bx}{n}

\providecommand{\norm}[1]{\|#1\|}

\newcommand\redout{\bgroup\markoverwith {\textcolor{red}{\rule[.5ex]{2pt}{0.4pt}}}\ULon}
\usepackage{xcolor}

\newcommand\model{SMART}

\begin{document}

\title{SMART: Robust and Efficient Fine-Tuning for Pre-trained Natural Language Models through Principled Regularized Optimization}

\author{Haoming Jiang, Pengcheng He, Weizhu Chen, Xiaodong Liu, Jianfeng Gao, Tuo Zhao \thanks{Work was done during Haoming Jiang's internship at Microsoft Dynamics 365 AI. Haoming Jiang and Tuo Zhao are affiliated with Georgia Institute of Technology. Pengcheng He and Weizhu Chen are affiliated with Microsoft Dynamics 365 AI. Xiaodong Liu and Jianfeng Gao are affiliated with Microsoft Research. Emails: \texttt{jianghm@gatech.edu}, \texttt{\{penhe,wzchen\}@microsoft.com}, \texttt{\{xiaodl,jfgao\}@microsoft.com}, \texttt{tourzhao@gatech.edu}. }}

\date{}

\maketitle


\begin{abstract}
\vspace{-0.05in}

Transfer learning has fundamentally changed the landscape of natural language processing (NLP). Many state-of-the-art models are first pre-trained on a large text corpus and then fine-tuned on downstream tasks. However, due to limited data resources from downstream tasks and the extremely high complexity of pre-trained models, aggressive fine-tuning often causes the fine-tuned model to overfit the training data of downstream tasks and fail to generalize to unseen data. 
To address such an issue in a principled manner, we propose a new learning framework for robust and efficient fine-tuning for pre-trained models to attain better generalization performance.
The proposed framework contains two important ingredients: 1. Smoothness-inducing regularization, which effectively manages the complexity of the model; 2.  Bregman proximal point optimization, which is an instance of trust-region methods and can prevent aggressive updating. 
Our experiments show that the proposed framework achieves new state-of-the-art performance on a number of NLP tasks including GLUE, SNLI, SciTail and ANLI. Moreover, it also outperforms the state-of-the-art T5 model, which is the largest pre-trained model containing 11 billion parameters, on GLUE.
\footnote{\url{https://github.com/namisan/mt-dnn}}

\end{abstract}



\section{Introduction}
\label{sec:intro}

The success of natural language processing (NLP) techniques relies on huge amounts of labeled data in many applications. However, large amounts of labeled data are usually prohibitive or expensive to obtain. To address this issue, researchers have resorted to transfer learning. 

Transfer learning considers the scenario, where we have limited labeled data from the target domain for a certain task, but we have relevant tasks with a large amount of data from different domains (also known as out-of-domain data). The goal is to transfer the knowledge from the high-resource domains to the low-resource target domain. Here we are particularly interested in the popular two-stage transfer learning framework \citep{pan2009survey}. The first stage is pre-training, where a high-capacity model is trained for the out-of-domain high-resource relevant tasks. The second stage is fine-tuning, where the high-capacity model is adapted to the low-resource task in the target domain.  

For many applications in NLP, most popular transfer learning methods choose to pre-train a large language model, e.g., ELMo \citep{peters2018deep}, GPT \citep{radford2019language} and BERT \citep{devlin2018bert}. 
Such a language model can capture general semantic and syntactic information that can be further used in downstream NLP tasks. 
The language model is particularly attractive, because it can be trained in a completely unsupervised manner with huge amount of unlabeled data, which are extremely cheap to fetch from internet nowadays. 
The resulting extremely large multi-domain text corpus allows us to train huge language models. To the best of our knowledge, by far the largest language model, T5, has an enormous size of about 11 billion parameters \citep{raffel2019t5}.


For the second fine-tuning stage, researchers adapt the pre-trained language model to the target task/domain. They usually replace the top layer of the language model by a task/domain-specific sub-network, and then continue to train the new model
with the limited data of the target task/domain. 
Such a fine-tuning approach accounts for the low-resource issue in the target task/domain, and has achieved state-of-the-art performance in many popular NLP benchmarks \citep{devlin2018bert,liu2019roberta,yang2019xlnet,lan2019albert,dong2019unified,raffel2019t5}. 

Due to the limited data from the target task/domain and the {\bf extremely high complexity} of the pre-trained model,  {\bf aggressive fine-tuning} often makes the adapted model overfit the training data of the target task/domain and therefore does not generalize well to unseen data. 
To mitigate this issue, the fine-tuning methods often rely on hyper-parameter tuning heuristics. For example, \citet{howard2018universal} use a heuristic learning rate schedule and gradually unfreeze the layers of the language model to improve the fine-tune performance; \citet{peters2019tune} give a different suggestion that they only adapt certain layers and freeze the others; \cite{houlsby2019parameter,stickland2019bert} propose to add additional layers to the pre-trained model and fine-tune both of them or only the additional layers. However, these methods require significant tuning efforts.

To fully harness the power of fine-tuning in a more principled manner, we propose a new learning framework for robust and efficient fine-tuning on the pre-trained language models through regularized optimization techniques. Specifically, our framework consists of two important ingredients for preventing overfitting: 

\noindent {\bf (I)} To effectively control the {\bf extremely high complexity} of the model, we propose a {\it Smoothness-inducing Adversarial Regularization} technique. Our proposed regularization is motivated by local shift sensitivity in existing literature on robust statistics. Such regularization encourages the output of the model not to change much, when injecting a small perturbation to the input. Therefore, it enforces the smoothness of the model, and effectively controls its capacity \citep{mohri2018foundations}.

\noindent {\bf (II)} To prevent {\bf aggressive updating}, we propose a class of {\it Bregman Proximal Point Optimization} methods. Our proposed optimization methods introduce a trust-region-type regularization \citep{conn2000trust} at each iteration, and then update the model only within a small neighborhood of the previous iterate. Therefore, they can effectively prevent aggressive updating and stabilize the fine-tuning process.

We compare our proposed method with several state-of-the-art competitors proposed in \cite{zhu2019freelb,liu2019mt-dnn,liu2019roberta,lan2019albert,raffel2019t5} and show that our proposed method significantly improves the training stability and generalization, and achieves comparable or better performance on multiple NLP tasks. We highlight that our single model with 356M parameters (without any ensemble) can achieve three state-of-the-art results on GLUE, even compared with all existing ensemble models and the T5 model \citep{raffel2019t5}, which contains 11 billion parameters. Furthermore, we also demonstrate that the proposed framework complements with SOTA fine-tuning methods \citep{liu2019mt-dnn} and outperforms the T5 model.


We summarize our contribution as follows:
1. We introduce the smoothness-inducing adversarial regularization and proximal point optimization into large scale language model fine-tuning;
2. We achieve state-of-the-art results on several popular NLP benchmarks (e.g., GLUE, SNLI, SciTail, and ANLI).

\noindent \textbf{Notation:} We use $f(x;\theta)$ to denote a mapping $f$ associated with the parameter $\theta$ from input sentences $x$ to an output space, where the output is a multi-dimensional probability simplex for classification tasks and a scalar for regression tasks. $\Pi_{\cA}$ denotes the projection operator to the set $\cA$. $\cD_{KL}(P||Q) = \sum_{k} p_k \log (p_k/q_k)$ denotes the KL-divergence of two discrete distributions $P$ and $Q$ with the associated parameters of $p_k$ and $q_k$, respectively.

\vspace{-0.05in}
\section{Background} \label{sec:background}
\vspace{-0.05in}


The transformer models were originally proposed in \citet{vaswani2017attention} for neural machine translation. Their superior performance motivated \citet{devlin2018bert} to propose a bidirectional transformer-based language model named BERT. Specifically, \citet{devlin2018bert} pre-trained the BERT model using a large corpus without any human annotation through unsupervised learning tasks. BERT motivated many follow-up works to further improve the pre-training by introducing new unsupervised learning tasks \citep{yang2019xlnet,dong2019unified,joshi2019spanbert}, enlarging model size \citep{lan2019albert,raffel2019t5}, enlarging training corpora \citep{liu2019roberta,yang2019xlnet,raffel2019t5} and  multi-tasking \citep{liu2019mt-dnn-kd,liu2019mt-dnn}.

The pre-trained language model is then adapted to downstream tasks and further fine-tuned. Specifically, the top layer of the language model can be replaced by a task-specific layer and then continue to train on downstream tasks.
To prevent overfitting, existing heuristics include choosing a small learning rate or a triangular learning rate schedule, and a small number of iterations, and other fine-tuning tricks mentioned in \cite{howard2018universal,peters2019tune,houlsby2019parameter,stickland2019bert}.

Our proposed regularization technique is related to several existing works \citep{miyato2018virtual,zhang2019theoretically,shu2018dirt}. These works consider similar regularization techniques, but target at other applications with different motivations, e.g., semi-supervised learning, unsupervised domain adaptation and harnessing adversarial examples in image classification.

Our proposed optimization technique covers a large  class of Bregman proximal point methods in existing literature on optimization, including vanilla proximal point method \citep{rockafellar1976monotone}, generalized proximal point method \citep{teboulle1997convergence,eckstein1993nonlinear}, accelerated proximal point method, and other variants \citep{guler1991convergence,guler1992new,parikh2014proximal}. 

There is a related fine-tuning method -- FreeLB \citep{zhu2019freelb}, which adapted a robust adversarial training method. However, our framework focuses on the local smoothness, leading to a significant performance improvement. More discussion and comparison are provided in Section \ref{sec:exp}. 
\vspace{-0.05in}
\section{The Proposed Method}
\label{sec:method}
\vspace{-0.05in}

We describe the proposed learning framework -- \textit{\textbf{SMART}} for robust and efficient fine-tuning of pre-trained language models. Our framework consists of two important ingredients: \textit{\textbf{SM}oothness-inducing \textbf{A}dversarial \textbf{R}egularization} and \textit{B\textbf{R}egman p\textbf{R}oximal poin\textbf{T} op\textbf{T}imization}\footnote{The complete name of our proposed method is \textit{\textbf{SMAR$^3$T$^2$}}, but we use \textit{\textbf{SMART}} for notational simplicity.}. 

\vspace{-0.05in}
\subsection{Smoothness-Inducing Adversarial Regularization}\label{sec:regularization}

We propose to impose an explicit regularization to effectively control the model complexity at the fine-tuning stage. Specifically, given the model $f(\cdot;\theta)$ and $n$ data points of the target task denoted by $\{(x_i,y_i)\}_{i=1}^n$, where $x_i$'s denote the embedding of the input sentences obtained from the first embedding layer of the language model and $y_i$'s are the associated labels, our method essentially solves the following optimization for fine-tuning:
\begin{align}
\textstyle\min_{\theta} \cF(\theta)=\cL(\theta) + \lambda_{\rm s} \cR_{\rm s}(\theta), 
\label{eq:smartobj}
\end{align}
where $\cL(\theta)$ is the loss function defined as
\begin{align*}
\textstyle\cL(\theta) = \frac{1}{n}\sum_{i=1}^n\ell(f(x_i;\theta),y_i),
\end{align*}
and $\ell(\cdot,\cdot)$ is the loss function depending on the target task, $\lambda_{\rm s}>0$ is a tuning parameter, and $\cR_{\rm s}(\theta)$ is the smoothness-inducing adversarial regularizer. Here we define $\cR_{\rm s}(\theta)$ as
\begin{align*}
\cR_{\rm s}(\theta) = \frac{1}{n}\sum_{i=1}^n \max_{\|\tilde{x}_i-x_i\|_p \leq \epsilon} \ell_{\rm s}(f(\tilde{x}_i;\theta),f(x_i;\theta)),
\end{align*}
where $\epsilon>0$ is a tuning parameter. Note that for classification tasks, $f(\cdot;\theta)$ outputs a probability simplex and $\ell_{\rm s}$ is chosen as the symmetrized KL-divergence, i.e., $$\ell_{\rm s}(P,Q) = \cD_{\rm KL}(P||Q)+\cD_{\rm KL}(Q||P);$$ For regression tasks, $f(\cdot;\theta)$ outputs a scalar and $\ell_{\rm s}$ is chosen as the squared loss, i.e., $\ell_s(p,q)=(p-q)^2$. Note that the computation of $\cR_{\rm s}(\theta)$ involves a maximization problem and can be solved efficiently by  projected gradient ascent. 

We remark that the proposed smoothness-inducing adversarial regularizer was first used in \citet{miyato2018virtual} for semi-supervised learning with $p=2$, and then in \citet{shu2018dirt} for unsupervised domain adaptation with $p=2$, and more recently in \citet{zhang2019theoretically} for harnessing the adversarial examples in image classification with $p=\infty$. To the best of our knowledge,
we are the first applying such a regularizer to fine-tuning of pre-trained language models.

\begin{figure}[htb!]
\centering
\includegraphics[width=0.6\linewidth]{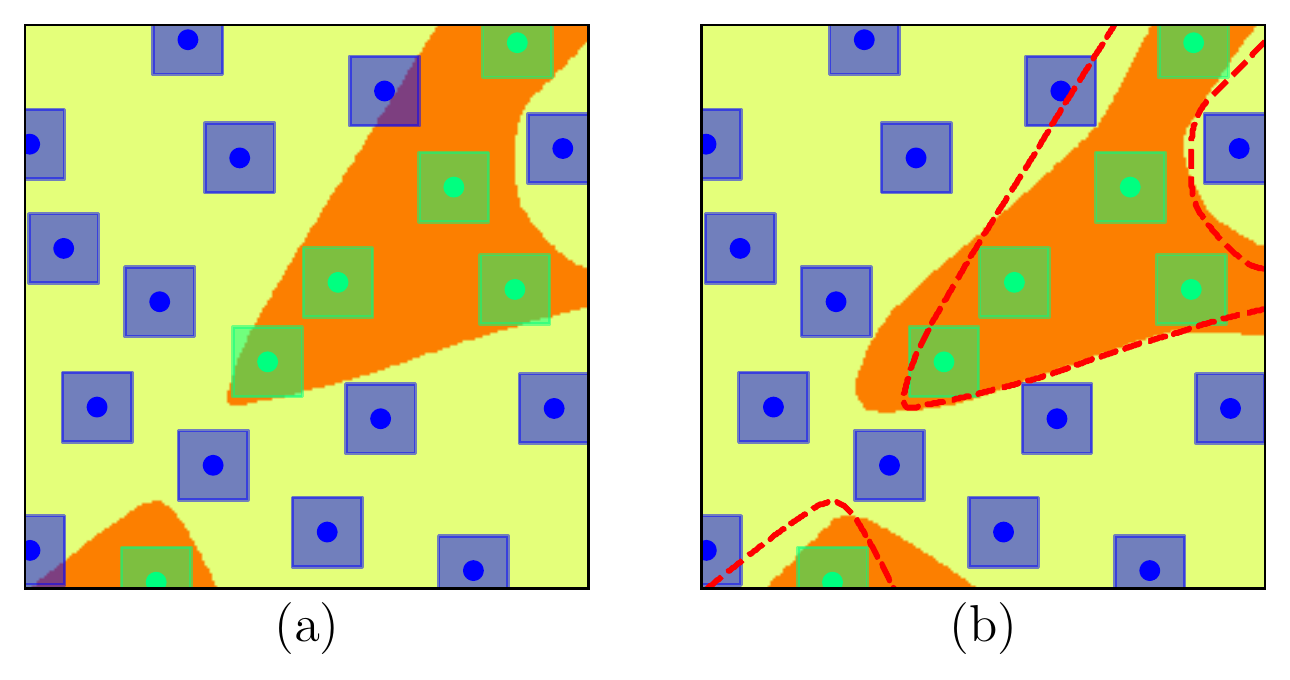}
\caption{\label{fig:TRADES} Decision boundaries learned without (a) and with (b) smoothness-inducing adversarial regularization, respectively. 
The red dotted line in (b) represents the decision boundary in (a). 
As can be seen, the output $f$ in (b) does not change much within the neighborhood of training data points.}
\vspace{-0.2in}
\end{figure}

The smoothness-inducing adversarial regularizer is essentially measuring the local Lipschitz continuity of $f$ under the metric $\ell_s$. More precisely speaking, the output of $f$ does not change much if we inject a small perturbation ($\ell_p$ norm bounded by $\epsilon$) to $x_i$. Therefore, by minimizing the objective in \eqref{eq:smartobj}, we can encourage $f$ to be smooth within the neighborhoods of all $x_i$'s. Such a smoothness-inducing property is particularly helpful to prevent overfitting and improve generalization on a low resource target domain for a certain task. An illustration is provided in Figure \ref{fig:TRADES}.

Note that the idea of measuring the local Lipschitz continuity is similar to the local shift sensitivity criterion in existing literature on robust statistics, which dates back to 1960's \citep{hampel1974influence,huber2011robust}. This criterion has been used to characterize the dependence of an estimator on the value of one of the sample points.

\subsection{Bregman Proximal Point Optimization}

We propose to develop a class of Bregman proximal point optimization methods to solve \eqref{eq:smartobj}. Such optimization methods impose a strong penalty at each iteration to prevent the model from aggressive update. Specifically, we use a pre-trained model as the initialization denoted by $f(\cdot;\theta_0)$. At the $(t+1)$-th iteration, the vanilla Bregman proximal point (VBPP) method takes
\begin{align}\label{eq:bpp_obj}
\textstyle\theta_{t+1} = \argmin_{\theta}\cF(\theta) + \mu\cD_{\rm Breg}(\theta,\theta_t),
\end{align}
where $\mu>0$ is a tuning parameter, and $\cD_{\rm Breg}(\cdot,\cdot)$ is the Bregman divergence defined as
\begin{align*}
\textstyle\cD_{\rm Breg}(\theta,\theta_t) = \frac{1}{n}\sum_{i=1}^n \ell_{\rm s}(f(x_i;\theta), f(x_i;\theta_t)),
\end{align*}
where $\ell_s$ is defined in Section \ref{sec:regularization}. As can be seen, when $\mu$ is large, the Bregman divergence at each iteration of the VBPP method essentially serves as a strong regularizer and prevents $\theta_{t+1}$ from deviating too much from the previous iterate $\theta_t$. This is also known as the trust-region type iteration in existing optimization literature \citep{conn2000trust}. Consequently, the Bregman proximal point method can effectively retain the knowledge of the out-of-domain data in the pre-trained model $f(\cdot;\theta_0)$. Since each subproblem \eqref{eq:bpp_obj} of VBPP does not admit a closed-form solution, we need to solve it using SGD-type algorithms such as ADAM. Note that we do not need to solve each subproblem until convergence. A small number of iterations are sufficient to output a reliable initial solution for solving the next subproblem.

Moreover, the Bregman proximal point method is capable of adapting to the information geometry (See more details in \citet{raskutti2015information}) of machine learning models and achieving better computational performance than the standard proximal point method (i.e., $\cD_{\rm Breg}(\theta,\theta_t) = \norm{\theta-\theta_t}_2^2$) in many applications.

%
\noindent \textbf{Acceleration by Momentum}. Similar to other optimization methods in existing literature, we can accelerate the Bregman proximal point method by introducing an additional momentum to the update. Specifically, at the $(t+1)$-th iteration, the momentum Bregman proximal point (MBPP) method takes
\begin{align}\label{eq:abpp_obj}
\textstyle\theta_{t+1} = \argmin_{\theta}\cF(\theta) + \mu\cD_{\rm Breg}(\theta,\tilde{\theta}_t),
\end{align}
where $\tilde{\theta}_t=(1-\beta)\theta_{t}+\beta\tilde{\theta}_{t-1}$ is the exponential moving average and $\beta\in(0,1)$ is the momentum parameter. The MBPP method is also called the ``Mean Teacher'' method in existing literature \citep{tarvainen2017mean} and has been shown to achieve state-of-the-art performance in popular semi-supervised learning benchmarks. For convenience, we summarize the  MBPP method in Algorithm \ref{algo:main}.

\begin{algorithm}[!tb]
\caption{SMART: We use the smoothness-inducing adversarial regularizer with $p=\infty$ and the momentum Bregman proximal point method.
}\label{algo:main}
\begin{algorithmic}[1]
    \item[\textbf{Notation:}] For simplicity, we denote $g_i(\tilde{x}_i,\bar{\theta}_s) =  \frac{1}{|\cB|}\sum_{x_i\in\cB}\nabla_{\tilde{x}} \ell_{\rm s}(f(x_i;\bar{\theta}_s),f(\tilde{x}_i;\bar{\theta}_s))$ and $\mathrm{AdamUpdate}_{\cB}$ denotes the ADAM update rule for optimizing \eqref{eq:abpp_obj} using the mini-batch $\cB$; $\Pi_{\cA}$ denotes the projection to $\cA$.
    
	\INPUT $T$: the total number of iterations, $\cX$: the dataset, $\theta_0$: the parameter of the pre-trained model, $S$: the total number of iteration for solving \eqref{eq:bpp_obj}, $\sigma^2$: the variance of the random initialization for $\tilde{x}_i$'s, $T_{\tilde{x}}$: the number of iterations for updating $\tilde{x}_i$'s, $\eta$: the learning rate for updating $\tilde{x}_i$'s, $\beta$: momentum parameter.

	\State $\tilde{\theta}_1 \leftarrow \theta_0$
	\For{$t=1,..,T$}
	\State $\bar{\theta}_1 \leftarrow \theta_{t-1}$
	\For{$s=1,..,S$}
	    \State Sample a mini-batch $\cB$ from $\cX$
	    \State For all $x_i\in\cB$, initialize $\tilde x_i \leftarrow x_i + \nu_i$ with $\nu_i\sim \cN(0,\sigma^2I)$
    	\For{$m=1,..,T_{\tilde{x}}$}
    	    \State $\tilde{g}_i \leftarrow \frac{g_i(\tilde{x}_i,\bar{\theta}_s)}{\norm{g_i(\tilde{x}_i,\bar{\theta}_s)}_{\infty}}$
    	    \State $\tilde x_i \leftarrow \Pi_{\|\tilde x_i-x\|_{\infty} \leq \epsilon} (\tilde x_i + \eta \tilde{g}_i)$ 
    	\EndFor
	    \State $\bar{\theta}_{s+1}\leftarrow {\rm Adam Update}_{\cB}(\bar{\theta}_{s})$  
	\EndFor
	\State $\theta_{t} \leftarrow \bar{\theta}_{S}$
	\State $\tilde{\theta}_{t+1}\leftarrow (1-\beta)\bar{\theta}_{S}+\beta\tilde{\theta}_{t}$
	\EndFor
		\OUTPUT $\theta_T$
\end{algorithmic}
\end{algorithm}


\vspace{-0.075in}
\section{Experiment -- Main Results}
\label{sec:exp}
\vspace{-0.05in}

We demonstrate the effectiveness of SMART for fine-tuning large language models using GLUE \cite{wang2018glue} by comparing with existing state-of-the-art methods. Dataset details can be found in Appendix \ref{app:dataset}.

\vspace{-0.1in}
\subsection{Implementation Details}
\label{subsec:impl}
\vspace{-0.05in}

Our implementation of {\model} is based on  BERT\footnote{https://github.com/huggingface/transformers} \citep{Wolf2019transfomerhf}, RoBERTa \footnote{https://github.com/pytorch/fairseq} \citep{liu2019roberta}, MT-DNN \footnote{https://github.com/namisan/mt-dnn} \citep{liu2020mtmtdnn} and HNN\footnote{https://github.com/namisan/mt-dnn/tree/master/hnn}.
We used ADAM \citep{kingma2014adam} and RADAM \citep{liu2019radam} as our optimizers with a learning rate in the range $\in \{1\times 10^{-5}, 2\times 10^{-5}, 3 \times 10^{-5}, 5\times 10^{-5}\}$ and a batch size $\in \{16, 32, 64\}$. 
The maximum number of epochs was set to $6$. 
A linear learning rate decay schedule with warm-up of $0.1$ was used, unless stated otherwise.
We also set the dropout rate of all the task specific layers as $0.1$, except $0.3$ for MNLI and $0.05$ for CoLA. 
To avoid gradient exploding, we clipped the gradient norm within $1$. 
All the texts were tokenized using wordpieces and were chopped to spans no longer than $512$ tokens. For {\model}, we set the perturbation size $\epsilon = 10^{-5}$ and $\sigma = 10^{-5}$. We set $\mu=1$ and $\lambda_s \in \{1,3,5\}$. The learning rate $\eta$ in Algorithm~\ref{algo:main} is set to $10^{-3}$.  We set $\beta$ = 0.99 for the first $10\%$ of the updates ($t\leq 0.1T$) and $\beta = 0.999$ for the rest of the updates  ($t>0.1T$) following \cite{tarvainen2017mean}. 
Lastly, we simply set $S=1,T_{\tilde x}=1$ in Algorithm~\ref{algo:main}. 

\subsection{GLUE Main Results}
\label{subsec:results}

We compare {\model} with a range of strong baselines including large pre-trained models and approaches with adversarial training, and a list of state-of-the-art models that have been submitted to the GLUE leaderboard. {\model} is a generic framework, we evaluate our framework on two pre-trained models, the BERT\textsubscript{BASE} model \citep{devlin2018bert} and the RoBERTa\textsubscript{LARGE} model \citep{liu2019roberta}, which are available publicly. Most of our analyses are done with the BERT\textsubscript{BASE} to make our results comparable to other work, since BERT\textsubscript{BASE} has been widely used as a baseline. To make our result comparable to other state-of-the-art models, we also evaluate the framework on the RoBERTa\textsubscript{LARGE} model. 

\vskip1pt
\noindent $\bullet$ BERT \citep{devlin2018bert}: This is the BERT\textsubscript{BASE} model released by the authors. In \citet{devlin2018bert}, authors only reported the development results on a few tasks, thus we reproduced the baseline results, which are denoted by \textbf{BERT\textsubscript{ReImp}}.  

\vskip1pt
\noindent $\bullet$ RoBERTa \citep{liu2019roberta}: This is the RoBERTa\textsubscript{LARGE} released by authors, and we present the reported results on the GLUE dev. 

\vskip1pt
\noindent $\bullet$ PGD, FreeAT, FreeLB \citep{zhu2019freelb}: They are three adversarial training approaches built on top of the RoBERTa\textsubscript{LARGE}. 




\vskip1pt
\noindent $\bullet$ {\model}: our proposed method as described in section~\ref{sec:method}. We use both the BERT\textsubscript{BASE} model ({\model}\textsubscript{BERT}) and the RoBERTa\textsubscript{LARGE} model ({\model}\textsubscript{RoBERTa}) as the pretrained model to evaluate the effectiveness of {\model}.  

\vskip1pt
The main results are reported in Table~\ref{tab:glue_dev}. This table can be clustered into two groups based on different pretrained models: the BERT\textsubscript{BASE} model (the first group) and the RoBERTa\textsubscript{LARGE} model (the second group). The detailed discussions are as follows.

For a fair comparison, we reproduced the BERT baseline (BERT\textsubscript{ReImp}), since several results on the GLUE development set were missed. Our reimplemented BERT baseline is even stronger than the originally reported results in \citet{devlin2018bert}. For instance, the reimplemented model obtains 84.5\% (vs. 84.4\%) on MNLI in-domain development in terms of accuracy. On SST-2, BERT\textsubscript{ReImp} outperforms BERT by 0.2\% (92.9\% vs. 92.7\%) accuracy. All these results demonstrate the fairness of our baselines.

\begin{table*}[!htb]
	\begin{center}
		\begin{tabular}{@{\hskip1pt}l@{\hskip1pt}|@{\hskip1pt}c@{\hskip1pt}|c@{\hskip1pt}|c@{\hskip1pt}|c@{\hskip1pt}|c@{\hskip1pt}|@{\hskip1pt}c @{\hskip1pt}|@{\hskip1pt} c @{\hskip1pt}|@{\hskip1pt}c@{\hskip1pt}}
			\hline \bf Model            &MNLI-{m/mm}        &QQP                &RTE            &QNLI           &MRPC               &CoLA           &SST          &STS-B\\ 
			                            &Acc                &Acc/F1             &Acc            &Acc            &Acc/F1             &Mcc            &Acc            &P/S Corr       \\ \hline
			\hline
            \multicolumn{9}{c}{\textbf{BERT\textsubscript{BASE}}}\\ \hline	
			BERT \citep{devlin2018bert}   &84.4/-             &-                  &-              &88.4           &-/86.7             &-              &92.7           &-\\ \hline
			BERT\textsubscript{ReImp}               &84.5/84.4          &90.9/88.3          &63.5           &91.1           &84.1/89.0          &54.7           &92.9           &89.2/88.8\\  \hline
            {\model}\textsubscript{BERT}                     &\textbf{85.6/86.0} &\textbf{91.5/88.5} & \textbf{71.2} & \textbf{91.7} &\textbf{87.7/91.3} & \textbf{59.1} &\textbf{93.0}  &\textbf{90.0/89.4} \\ \hline \hline
        \multicolumn{9}{c}{\textbf{RoBERTa\textsubscript{LARGE}}}\\	 \hline		
        RoBERTa \citep{liu2019roberta}    &90.2/-         &92.2/-                 &86.6           &94.7           &-/90.9             &68.0           &96.4           &92.4/-\\ \hline 
        PGD \citep{zhu2019freelb}         &90.5/-         &92.5/-                 &87.4           &94.9           &-/90.9             &69.7           &96.4           &92.4/- \\ \hline
        FreeAT \citep{zhu2019freelb}      &90.0/-         &92.5/-                 &86.7           &94.7           &-/90.7             &68.8           &96.1           &92.4/- \\ \hline
        FreeLB \citep{zhu2019freelb}      &90.6/-         &\textbf{92.6}/-        &88.1           &95.0           &-/91.4             &\textbf{71.1}  &96.7           &92.7/- \\ \hline \hline
        {\model}\textsubscript{RoBERTa} &\textbf{91.1/91.3}  &92.4/89.8         &\textbf{92.0}  &\textbf{95.6}  &\textbf{89.2/92.1} &70.6           &\textbf{96.9}  &\textbf{92.8/92.6} \\ \hline
		\end{tabular}
	\end{center}
	\caption{Main results on GLUE development set. The best result on each task produced by a single model is in \textbf{bold} and ``-'' denotes the missed result.
	}
	\label{tab:glue_dev}
\end{table*}
\begin{table*}[htb!]
\scriptsize
	\begin{center}
		\begin{tabular}{@{\hskip1pt}l@{\hskip1pt}|@{\hskip1pt}l@{\hskip1pt}|@{\hskip1pt}c@{\hskip1pt}|@{\hskip1pt}c@{\hskip1pt}|@{\hskip1pt}c@{\hskip1pt}|@{\hskip1pt}c@{\hskip1pt}|@{\hskip1pt}c|@{\hskip1pt}c|@{\hskip1pt}c |@{\hskip1pt} c |@{\hskip1pt} c|@{\hskip1pt} c |@{\hskip1pt} c@{\hskip1pt}}
			\hline \bf Model /\#Train           &CoLA   &SST  &MRPC       &STS-B      &QQP        &MNLI-m/mm      &QNLI   &RTE    &WNLI   &AX     &\textbf{Score} &\#param\\
		& 8.5k &67k &3.7k &7k &364k &393k &108k &2.5k &634 & & \\ \hline \hline			
            Human Performance           &66.4   &97.8   &86.3/80.8  &92.7/92.6  &59.5/80.4  &92.0/92.8	    &91.2   &93.6   &95.9	&-      &87.1 &- \\ \hline \hline
            \multicolumn{13}{c}{\textbf{Ensemble Models}}\\ \hline	
            RoBERTa$^1$                 &67.8   &96.7   &92.3/89.8  &92.2/91.9  &74.3/90.2  &90.8/90.2	    &98.9   &88.2   &89.0	&48.7   &88.5 &356M\\ \hline
            FreeLB$^2$                  &68.0   &96.8   &93.1/90.8  &92.4/92.2  &\textbf{74.8}/90.3  &91.1/90.7	    &98.8   &88.7   &89.0	&50.1   &88.8 &356M \\ \hline
            ALICE$^3$                   &69.2   &97.1   &93.6/91.5  &92.7/92.3  &74.4/\textbf{90.7}  &90.7/90.2	    &\textbf{99.2}   &87.3   &89.7	&47.8   &89.0 &340M\\ \hline
            ALBERT$^4$                  &69.1   &97.1   &93.4/91.2  &92.5/92.0  &74.2/90.5  &91.3/91.0	    &\textbf{99.2}   &89.2   &91.8	&50.2   &89.4 &235M$^*$\\ \hline 
            MT-DNN-{\model}$^\dagger$                   &69.5   &\textbf{97.5}   &\textbf{93.7/91.6}  &\textbf{92.9/92.5}  &73.9/90.2  &91.0/90.8	    &\textbf{99.2}   &89.7   & 94.5	& 50.2  & \textbf{89.9} & 356M\\ \hline \hline
            \multicolumn{13}{c}{\textbf{Single Model}}\\ \hline	
			BERT$_{\text{LARGE}}$$^5$   &60.5   &94.9   &89.3/85.4  &87.6/86.5  &72.1/89.3  &86.7/85.9      &92.7   &70.1   &65.1	&39.6   &80.5 &335M\\ \hline
            MT-DNN$^6$                  &62.5   &95.6   &90.0/86.7  &88.3/87.7  &72.4/89.6  &86.7/86.0	    &93.1   &75.5   &65.1	&40.3   &82.7 &335M\\ \hline
            T5$^8$                          &\textbf{70.8}   &97.1   &91.9/89.2  &92.5/92.1  &74.6/90.4  &\textbf{92.0/91.7}	    &96.7   &\textbf{92.5}   &\textbf{93.2}	&\textbf{53.1}   &89.7 &11,000M\\ \hline
            {\model}\textsubscript{RoBERTa}                   &65.1   &\textbf{97.5}   &\textbf{93.7/91.6}  &\textbf{92.9/92.5}  &74.0/90.1  &91.0/90.8	    &95.4   &87.9   & 91.8$^8$	& 50.2  & 88.4 & 356M\\ \hline
		\end{tabular}
	\end{center}
	\caption{GLUE test set results scored using the GLUE evaluation server. 
	The state-of-the-art results are in \textbf{bold}.
	All the results were obtained from \href{https://gluebenchmark.com/leaderboard}{https://gluebenchmark.com/leaderboard} on December 5, 2019.
	{\model} uses the classification objective on QNLI. Model references: 
	$^1$ \protect{\citet{liu2019roberta}}; $^2$\protect{\citet{zhu2019freelb}}; $^3$\protect{\citet{wang2019alice}}; $^4$\protect{\citet{lan2019albert}}; 
	$^5$ \protect{\citet{devlin2018bert}};
	$^6$ \protect{\citet{liu2019mt-dnn}};
	$^7$ \protect{\citet{raffel2019t5}} 
	and $^8$ \citet{he2019hnn}, \citet{kocijan2019surprisingly}. $^*$ ALBERT uses a model similar in size, architecture and computation cost to a 3,000M BERT (though it has dramatically fewer parameters due to parameter sharing).  $^\dagger$ Mixed results from ensemble and single of MT-DNN-{\model} and with data augmentation.
	} 
	\label{tab:glue_test}
\vspace{-0.1in}
\end{table*}

Comparing with two strong baselines BERT and RoBERTa \footnote{In our experiments, we use BERT referring the BERT\textsubscript{BASE} model, which has 110 million parameters, and RoBERTa referring the RoBERTa\textsubscript{LARGE} model, which has 356 million parameters, unless stated otherwise.}, {\model}, including {\model}\textsubscript{BERT} and {\model\textsubscript{RoBERTa}}, consistently outperforms them across all 8 GLUE tasks by a big margin. 
Comparing with BERT, {\model}\textsubscript{BERT} obtained 85.6\% (vs. 84.5\%) and 86.0\% (vs. 84.4\%) in terms of accuracy, which is 1.1\% and 1.6\% absolute improvement, on the MNLI in-domain and out-domain settings. Even comparing with the state-of-the-art model RoBERTa, {\model}\textsubscript{RoBERTa} improves 0.8\% (91.1\% vs. 90.2\%) on MNLI in-domain development set. Interestingly, on the MNLI task, the performance of {\model} on the out-domain setting is better than the in-domain setting, e.g., (86.0\% vs. 85.6\%) by {\model}\textsubscript{BERT} and (91.3\% vs. 91.1\%) by {\model}\textsubscript{RoBERTa}, showing that our proposed approach alleviates the domain shifting issue. Furthermore, on the small tasks, the improvement of {\model} is even larger. For example, comparing with BERT, {\model\textsubscript{BERT}} obtains 71.2\% (vs. 63.5\%) on RTE and 59.1\% (vs. 54.7\%) on CoLA in terms of accuracy, which are 7.7\% and 4.4\% absolute improvement for RTE and CoLA, respectively; similarly, {\model\textsubscript{RoBERTa}} outperforms RoBERTa 5.4\% (92.0\% vs. 86.6\%) on RTE and 2.6\% (70.6\% vs. 68.0\%) on CoLA.    

We also compare {\model} with a range of models which used adversarial training such as FreeLB. From the bottom rows in Table~\ref{tab:glue_dev}, {\model} outperforms PGD and FreeAT across the all 8 GLUE tasks. Comparing with the current state-of-the-art adversarial training model, FreeLB, {\model} outperforms it on 6 GLUE tasks out of a total of 8 tasks (MNLI, RTE, QNLI, MRPC, SST-2 and STS-B) showing the effectiveness of our model. 

Table~\ref{tab:glue_test} summarizes the current state-of-the-art models on the GLUE leaderboard. {\model} obtains a competitive result comparing with T5 \citep{raffel2019t5}, which is the leading model at the GLUE leaderboard. T5 has 11 billion parameters, while {\model} only has 356 millions. Among this super large model (T5) and other ensemble models (e.g., ALBERT, ALICE), {\model}, which is a single model, still sets new state-of-the-art results on SST-2, MRPC and STS-B. By combining with the Multi-task Learning framework (MT-DNN), MT-DNN-{\model} obtains new state-of-the-art on GLUE, pushing the GLUE benchmark to 89.9\%. More discussion will be provided in Section \ref{subsec:smart-mtl}.

\vspace{-0.05in}
\section{Experiment -- Analysis and Extension}
\label{sec:exp_analysis}
\vspace{-0.05in}

In this section, we first analyze the effectiveness of each component of the proposed method. We also study that whether the proposed method is complimentary to multi-task learning. We further extend SMART to domain adaptation and use both SNLI \citep{snli2015} and SciTail \citep{scitail} to evaluate the effectiveness. Finally, we verified the robustness of the proposed method on ANLI \citep{nie2019adversarial}. 

\subsection{Ablation Study}
\label{subsec:abl}

Note that due to the limitation of time and computational resources, all the experiments reported below are based on the \textbf{BERT\textsubscript{BASE}} model.
In this section, we study the importance of each component of {\model}: smoothness-inducing adversarial regularization and Bregman proximal point optimization. All models in this study used the BERT\textsubscript{BASE} as the encoder for fast training. Furthermore, we also include the BERT\textsubscript{BASE} model as an additional baseline for a fair comparison. {\model} denotes the proposed model. Then we set $\lambda_s$ to 0, which denotes as -$\cR_{\rm s}$. The model with $\mu=0$ is noted as -$\cD_{\rm Breg}$.

\begin{table}[!htb]
    \centering
    \begin{tabular}{l|c@{\hskip1pt}|c@{\hskip1pt}|c@{\hskip1pt}|c@{\hskip1pt} | c@{\hskip1pt}}
    \hline
   Model                        &MNLI   &RTE    & QNLI  &SST    &MRPC \\ 
                                &Acc    &Acc    &Acc    &Acc    &Acc  \\\hline \hline 
      BERT                      &84.5   &63.5   &91.1   &92.9   &89.0 \\ \hline
     {\model}                   &\textbf{85.6}  &\textbf{71.2}  &\textbf{91.7}   &\textbf{93.0}   &\textbf{91.3} \\ \hline
     -$\cR_{\rm s}$             &84.8   &70.8   &91.3   &92.8   &90.8\\ \hline     
 -$\cD_{\rm Breg}$                &85.4   &\textbf{71.2}   &91.6   &92.9   &91.2\\ \hline     
    \end{tabular}
    \caption{Ablation study of {\model} on 5 GLUE tasks. Note that all models used the BERT\textsubscript{BASE} model as their encoder.}
    \label{tab:smart_abl}
\end{table}

The results are reported in Table~\ref{tab:smart_abl}. It is expected that the removal of either component (smooth regularization or proximal point method) in {\model} would result in a performance drop. For example, on MNLI, removing smooth regularization leads to a 0.8\% (85.6\% vs. 84.8) performance drop, while removing the Breg proximal point optimization, results in a performance drop of 0.2\% (85.6\% vs. 85.4\%). It demonstrates that these two components complement each other. Interestingly, all three proposed models outperform the BERT baseline model demonstrating the effectiveness of each module. Moreover, we obersere that the generalization performance benefits more from {\model}  on small datasets (i.e., RTE and MRPC) by preventing overfitting.

\subsection{Error Analysis}
\label{subsec:error}
To understand why {\model} improves the performance, we analyze it on the ambiguous samples of MNLI dev set containing 3 classes, where each sample has 5 annotations. Based on the degree of agreement between these annotations, we divide the samples into 4 categories: 
1) \textbf{5/0/0} all five annotations are the same; 2) \textbf{4/1/0} four annotations are the same; 3) \textbf{3/2/0} three annotations are the same and the other two annotations are the same; 4) \textbf{3/1/1} three annotations are the same and the other two annotations are different.

\begin{figure}[!htb]
    \centering
    \includegraphics[width=0.7\linewidth]{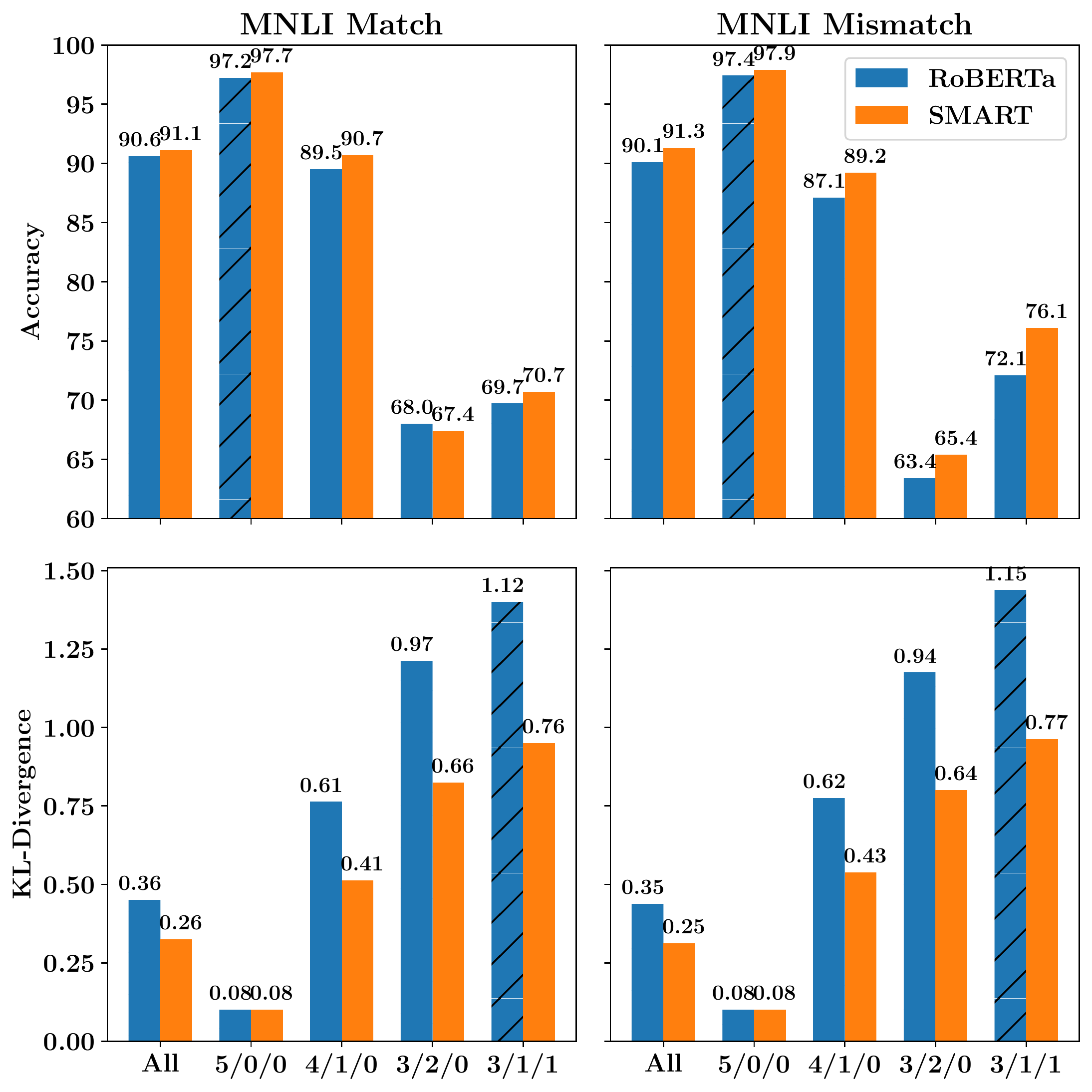}
    \caption{Score breakdown by degree of agreement.}
    \label{fig:mnli_ambi}
\end{figure}

Figure~\ref{fig:mnli_ambi} summarizes the results in terms of both accuracy and KL-divergence: $$-\frac{1}{n}\sum_{i=1}^n \sum_{j=1}^3 p_j(x_i)\log(f_j(x_i)).$$ For a given sample $x_i$, the KL-Divergence evaluates the similarity between the model prediction $\{f_j(x_i)\}_{j=1}^3$ and the annotation distribution $\{p_j(x_i)\}_{j=1}^3$. We observe that {\model}\textsubscript{RoBERTa} outperforms RoBERTa across all the settings. Further, on high degree of ambiguity (low degree of agreement), {\model}\textsubscript{RoBERTa} obtains an even larger improvement showing its robustness to ambiguity.


\subsection{{\model} with Multi-task Learning}
\label{subsec:smart-mtl}

It has been shown that multi-task learning (MTL, \citet{caruana1997multitask,liu2015mtl,liu2019mt-dnn}) has a regularization effect via alleviating overfitting to a specific task. One question is whether MTL helps {\model} as well. In this section, we are going to answer this question. Following \citet{liu2019mt-dnn}, we first ``pre-trained" shared embeddings using MTL with {\model}, denoted as \textbf{MT-DNN-{\model}} \footnote{Due to limitation of computational resources, we only trained jointly using MTL on MNLI, RTE, QNLI, SST and MRPC, while MT-DNN was trained on the whole GLUE tasks except CoLA. 
}, and then adapted the training data on each task on top of the shared embeddings. We also include a baseline which fine-tuned each task on the publicly released MT-DNN checkpoint \footnote{It is from: https://github.com/namisan/mt-dnn. Note that we did not use the complicated answer module, e.g., SAN \citep{liu2018san4nli}.}, which is indicated as \textbf{MT-DNN-{\model}\textsubscript{v0}}.

\begin{table}[!htb]
    \centering
    \begin{tabular}{@{\hskip1pt}l@{\hskip1pt}|@{\hskip1pt}c@{\hskip1pt}|@{\hskip1pt}c@{\hskip1pt}|@{\hskip1pt}c@{\hskip1pt}|@{\hskip1pt}c@{\hskip1pt} |@{\hskip1pt} c@{\hskip1pt}}
    \hline
   Model                        &MNLI   &RTE    & QNLI  &SST    &MRPC \\ 
                                &Acc    &Acc    &Acc    &Acc    &F1  \\\hline  \hline
     BERT                       &84.5   &63.5   &91.1   &92.9   &89.0 \\ \hline
     MT-DNN                     &85.3   &79.1   &91.5   &\textbf{93.6}   &89.2 \\ \hline 
     {\model}                   &85.6   &71.2   &91.6   &93.0   &91.3 \\ \hline
     MT-DNN-{\model}\textsubscript{v0}            &\textbf{85.7}   &80.2   &\textbf{92.0}   &93.3   &91.5 \\ \hline
     MT-DNN-{\model}            &\textbf{85.7}       &\textbf{81.2}   &\textbf{92.0}     &93.5       &\textbf{91.7}  \\ \hline
    \end{tabular}
    \caption{Comparison between {\model} and MTL. }
    \label{tab:smart_mtl}
\end{table}

We observe that both MT-DNN and {\model} consistently outperform the BERT model on all five GLUE tasks. Furthermore, {\model} outperforms MT-DNN on MNLI, QNLI, and MRPC, while it obtains worse results on RTE and SST, showing that MT-DNN is a strong counterpart for {\model}. By combining these two models, MT-DNN-{\model}\textsubscript{v0} enjoys advantages of both and thus improved the final results. For example, it achieves 85.7\% (+0.1\%) on MNLI and 80.2\% (+1.1\%) on RTE comparing with the best results of MT-DNN and {\model} demonstrating that these two techniques are orthogonal. Lastly we also trained {\model} jointly and then finetuned on each task like \citet{liu2019mt-dnn}. We observe that MT-DNN-{\model} outperformes MT-DNN-{\model}\textsubscript{v0} and MT-DNN across all 5 tasks (except MT-DNN on SST) showing that {\model} improves the generalization of MTL.

\subsection{Domain Adaptation}
\label{subsec:domain}

In this section, we evaluate our model on the domain adaptation setting. Following \citet{liu2019mt-dnn}, we start with the default training/dev/test set of SNLI and SciTail. Then, we randomly sample 0.1\%, 1\%, 10\% and 100\% of its training data, which is used to train a model. 

\begin{table}[tb!]
	\begin{center}
		\begin{tabular}{@{\hskip1pt}l@{\hskip1pt} |@{\hskip1pt} c @{\hskip1pt}|@{\hskip1pt} c @{\hskip1pt}|@{\hskip1pt} c @{\hskip1pt}|@{\hskip1pt} c@{\hskip1pt}}
			\hline \bf Model & 0.1\% & 1\% &10\% & 100\% \\ \hline
            \multicolumn{5}{c}{ SNLI Dataset (Dev Accuracy\%)} \\ \hline
            \#Training Data &549& 5,493& 54,936&549,367 \\ \hline
            BERT &52.5&78.1&86.7 & 91.0 \\ \hline
            MT-DNN &82.1 & 85.2 & 88.4 & 91.5 \\ \hline
            MT-DNN-{\model} &\textbf{82.7} & \textbf{86.0} & \textbf{88.7} &\textbf{91.6}  \\ \hline \hline

\multicolumn{5}{c}{ SciTail Dataset (Dev Accuracy\%)} \\ \hline
            \#Training Data &23& 235& 2,359& 23,596\\ \hline
            BERT &51.2&82.2&90.5 & 94.3 \\ \hline
            MT-DNN &81.9 & 88.3 & 91.1 & 95.8 \\ \hline			
            MT-DNN-{\model} &\textbf{82.3} &\textbf{88.6}  &\textbf{91.3} &\textbf{96.1}  \\ \hline
		\end{tabular}
	\end{center}
	\caption{Domain adaptation on SNLI and SciTail. 
	}
	\label{tab:domain}
\end{table}

The results are reported in Table~\ref{tab:domain}. We observe that both MT-DNN and MT-DNN-{\model} significantly outperform the BERT baseline. Comparing with MT-DNN, MT-DNN-{\model} also achieves some improvements indicating the robustness of {\model}. Furthermore, MT-DNN-{\model} outperforms current state-of-the-art on the SNLI/SciTail test. 

\begin{table*}[htb!]
    \centering
    \begin{tabular}{c|c|c|c|c|c|c|c|c}
    \hline
         \multirow{2}{*}{Method} & \multicolumn{4}{c|}{Dev} & \multicolumn{4}{c}{Test}  \\
         \cline{2-9}
          &  R1 & R2 & R3 & All & R1 & R2 & R3 & All  \\
          \hline 
 		\multicolumn{9}{c}{ MNLI + SNLI + ANLI + FEVER  }  \\ \hline
          BERT\textsubscript{LARGE}  \citep{nie2019adversarial}  &-&-&-& - &57.4&48.3&43.5& 49.3 \\
		\hline
		XLNet\textsubscript{LARGE} \citep{nie2019adversarial}  &-&-&-& -  &67.6&50.7&48.3& 55.1\\
		\hline
		RoBERTa\textsubscript{LARGE} \citep{nie2019adversarial}  &-&-&-& - &73.8&48.9&44.4& 53.7\\
		\hline 
		SMART\textsubscript{RoBERTa-LARGE}  & 74.5&50.9&47.6& \textbf{57.1} &72.4&49.8&50.3& \textbf{57.1} \\ \hline \hline
 		\multicolumn{9}{c}{ ANLI }  \\ 
 		\hline
 		RoBERTa\textsubscript{LARGE} \citep{nie2019adversarial} &-&-&-& -  & 71.3&43.3&43.0& 51.9 \\ 		\hline 
 		SMART\textsubscript{RoBERTa-LARGE}  &74.2&49.5&49.2& \textbf{57.1} &72.4&50.3&49.5& \textbf{56.9}\\
    \hline
    \end{tabular}
    \caption{Experiment Result for Each Round of ANLI.}
    \label{tab:anli_full}
\end{table*}
\begin{table}[!htb]
	\begin{center}
		\begin{tabular}{@{\hskip1pt}l @{\hskip1pt}|@{\hskip1pt} c @{\hskip1pt}|@{\hskip1pt} c@{\hskip1pt}}\hline
	   \bf Model &Dev& Test  \\ \hline 

		\multicolumn{3}{c}{ SNLI Dataset (Accuracy\%)}  \\ \hline 
		BERT\textsubscript{BASE}&91.0 & 90.8 \\ \hline		
		BERT\textsubscript{BASE}+SRL\citep{bertdep2019} &- & 90.3 \\ \hline		
		MT-DNN\textsubscript{BASE} &91.4 & 91.1 \\ \hline
		{\model}\textsubscript{BERT-BASE} &91.4& 91.1 \\ \hline
		MT-DNN-{\model}\textsubscript{BASE}\textsubscript{v0} &91.7& 91.4 \\ \hline
		MT-DNN-{\model}\textsubscript{BASE} &91.7&91.5 \\ \hline
		\hline
		BERT\textsubscript{LARGE}+SRL\citep{bertdep2019} & -& 91.3 \\ \hline
		BERT\textsubscript{LARGE} &91.7&91.0\\ \hline
		{MT-DNN\textsubscript{LARGE}} &92.2& 91.6\\ \hline
		{MT-DNN-{\model}\textsubscript{LARGE}}\textsubscript{v0} &\textbf{92.6}&\textbf{91.7}\\ \hline

		\hline
		\multicolumn{3}{c}{ SciTail Dataset (Accuracy\%)}  \\ \hline 	
		GPT \citep{gpt22019} &- &88.3 \\ \hline
		BERT\textsubscript{BASE} &94.3 & 92.0 \\ \hline
		MT-DNN\textsubscript{BASE} &95.8 &94.1 \\ \hline
		{\model}\textsubscript{BERT-BASE} &94.8& 93.2 \\ \hline
		MT-DNN-{\model}\textsubscript{BASE}\textsubscript{v0} &96.0&94.0  \\ \hline  
MT-DNN-{\model}\textsubscript{BASE} &96.1& 94.2 \\ \hline  
		\hline
		BERT\textsubscript{LARGE} &95.7& 94.4\\ \hline
		{MT-DNN}\textsubscript{LARGE} &96.3& 95.0\\ \hline

		\hline
{\model}\textsubscript{BERT-LARGE}& 96.2& 94.7\\
		\hline
MT-DNN-{\model}\textsubscript{LARGE}\textsubscript{v0} &\textbf{96.6}& \textbf{95.2}\\ \hline
		\end{tabular}
	\end{center}
\vspace{-0.125in}
\caption{Results on the SNLI and SciTail dataset.}	
	\label{tab:nli}
\end{table}

\subsection{Results on SNLI and SciTail}

In Table~\ref{tab:nli}, we compare our methods, using all in-domain training data, against several state-of-the-art models. We observe that {\model} obtains the same improvement on SNLI in the BERT setting. Combining {\model} with MT-DNN achieves a significant improvement, e.g., our BASE model even outperforms the BERT\textsubscript{LARGE} model. Similar observation is found on SciTail and in the BERT\textsubscript{LARGE} model setting. We see that incorporating {\model} into MT-DNN achieves new state-of-the-art results on both SNLI and SciTail, pushing benchmarks to 91.7\% on SNLI and 95.2\% on SciTail.  

\subsection{Robustness}

One important property of the machine learning model is its robustness to adversarial attack. We test our model on an adversarial natural language inference (ANLI) dataset \cite{nie2019adversarial}. 

We evaluate the performance of SMART on each subset (i.e., R1,R2,R3) of ANLI dev and test set. The results are presented in Table~\ref{tab:anli_full}. Table~\ref{tab:anli_full} shows the results of training on combined NLI data: ANLI \citep{nie2019adversarial} + MNLI \citep{mnli2018} + SNLI \cite{snli2015} + FEVER \citep{thorne2018fever} and training on only ANLI data. In the combined data setting, we obverse that 	{\model}\textsubscript{RoBERTa-LARGE} obtains the best performance compared with all the strong baselines, pushing benchmarks to 57.1\%. In case of the RoBERTa\textsubscript{LARGE} baseline,  {\model}\textsubscript{RoBERTa-LARGE} outperforms 3.4\% absolute improvement on dev and 7.4\% absolute improvement on test, indicating the robustness of {\model}. We obverse that in the ANLI-only setting, {\model}\textsubscript{RoBERTa-LARGE} outperforms the strong RoBERTa\textsubscript{LARGE} baseline with a large margin, +5.2\% (57.1\% vs. 51.9\%)
\section{Conclusion}
\label{sec:conclusion}
\vspace{-0.05in}

We propose a robust and efficient computation framework, {\model}, for fine-tuning large scale pre-trained natural language models in a principled manner. The framework effectively alleviates the overfitting and aggressive updating issues in the fine-tuning stage. 
{\model} includes two important ingredients: 1) smooth-inducing adversarial regularization; 2) Bregman proximal point optimization. 
Our empirical results suggest that {\model} improves the performance on many NLP benchmarks (e.g., GLUE, SNLI, SciTail, ANLI) with the state-of-the-art pre-trained models (e.g., BERT, MT-DNN, RoBERTa). We also demonstrate that the proposed framework is applicable to domain adaptation and results in a significant performance improvement. 
Our proposed fine-tuning framework can be generalized to solve other transfer learning problems. We will explore this direction as future work. 
\vspace{-3mm}
\section*{Acknowledgments}
\vspace{-3mm}
We thank Jade Huang, Niao He, Chris Meek, Liyuan Liu, Yangfeng Ji, Pengchuan Zhang, Oleksandr Polozov, Chenguang Zhu and Keivn Duh for valuable discussions and comments, and Microsoft Research Technology Engineering team for setting up GPU machines. We also thank the anonymous reviewers for valuable discussions. 

\bibliographystyle{ims}
\bibliography{ref}

\clearpage


\section{Datasets}
\label{app:dataset}

\begin{table*}[!htb]
	\begin{center}
		\begin{tabular}{l|l|c|c|c|c|c}
			\hline \bf Corpus &Task& \#Train & \#Dev & \#Test   & \#Label &Metrics\\ \hline \hline
			\multicolumn{6}{@{\hskip1pt}r@{\hskip1pt}}{Single-Sentence Classification (GLUE)} \\ \hline
			CoLA & Acceptability&8.5k & 1k & 1k & 2 & Matthews corr\\ \hline
			SST & Sentiment&67k & 872 & 1.8k & 2 & Accuracy\\ \hline \hline
			\multicolumn{6}{@{\hskip1pt}r@{\hskip1pt}}{Pairwise Text Classification (GLUE)} \\ \hline
			MNLI & NLI& 393k& 20k & 20k& 3 & Accuracy\\ \hline
            RTE & NLI &2.5k & 276 & 3k & 2 & Accuracy \\ \hline
            WNLI & NLI &634& 71& 146& 2 & Accuracy \\ \hline
			QQP & Paraphrase&364k & 40k & 391k& 2 & Accuracy/F1\\ \hline
            MRPC & Paraphrase &3.7k & 408 & 1.7k& 2&Accuracy/F1\\ \hline
			QNLI & QA/NLI& 108k &5.7k&5.7k&2& Accuracy\\ \hline \hline
			\multicolumn{5}{@{\hskip1pt}r@{\hskip1pt}}{Text Similarity (GLUE)} \\ \hline
			STS-B & Similarity &7k &1.5k& 1.4k &1 & Pearson/Spearman corr\\ \hline
			\multicolumn{6}{@{\hskip1pt}r@{\hskip1pt}}{Pairwise Text Classification} \\ \hline
			SNLI & NLI& 549k &9.8k&9.8k&3& Accuracy\\ \hline
			SciTail & NLI& 23.5k &1.3k&2.1k&2& Accuracy\\ \hline
			ANLI & NLI& 163k &3.2k&3.2k&3& Accuracy\\ \hline

		\end{tabular}
	\end{center}
	\caption{Summary of the four benchmarks: GLUE, SNLI, SciTail and ANLI.
	}
	\label{tab:datasets}
\end{table*}

The GLUE benchmark, SNLI, SciTail and ANLI is briefly introduced in the following sections. The detailed description can be found in \cite{wang2018glue, snli2015, scitail, nie2019adversarial}. Table~\ref{tab:datasets} summarizes the information of these tasks. 

\noindent $\bullet$ \textbf{GLUE}. The General Language Understanding Evaluation (GLUE) benchmark is a collection of nine natural language understanding (NLU) tasks. As shown in Table~\ref{tab:datasets},
it includes question answering~\cite{squad1}, linguistic acceptability~\cite{cola2018}, sentiment analysis~\cite{sst2013}, text similarity~\cite{sts-b2017}, paraphrase detection~\cite{mrpc2005}, and natural language inference (NLI)~\cite{rte1,rte2,rte3,rte5,winograd2012,mnli2018}. The diversity of the tasks makes GLUE very suitable for evaluating the generalization and robustness of NLU models. 

\noindent $\bullet$ \textbf{SNLI}.
The Stanford Natural Language Inference (SNLI) dataset contains 570k human annotated sentence pairs, in which the premises are drawn from the captions of the Flickr30 corpus and hypotheses are manually annotated \cite{snli2015}. 
This is the most widely used entailment dataset for NLI.
The dataset is used only for domain adaptation in this study.

\noindent $\bullet$ \textbf{SciTail}
This is a textual entailment dataset derived from a science question answering (SciQ) dataset \cite{scitail}. The task involves assessing whether a given premise entails a given hypothesis.  
In contrast to other entailment datasets mentioned previously, the hypotheses in SciTail are created from science questions while the corresponding answer candidates and premises come from relevant web sentences retrieved from a large corpus. As a result, these sentences are linguistically challenging and the lexical similarity of premise and hypothesis is often high, thus making SciTail particularly difficult. 
The dataset is used only for domain adaptation in this study.

\noindent $\bullet$ \textbf{ANLI}.
The Adversarial Natural Language Inference (ANLI, \citet{nie2019adversarial}) is a new large-scale NLI benchmark dataset, collected via an iterative, adversarial human-and-model-in-the-loop procedure. Particular, the data is selected to be difficult to the state-of-the-art models, including BERT and RoBERTa.

\section{Hyperparameters}
As for the sensitivities of hyper-parameters, in general the performance of our method is not very sensitive to the choice of hyper-parameters as detailed below. 
\begin{itemize}
    \item We only observed slight differences in model performance when $\lambda_s \in [1,10]$, $ \mu \in [1,10]$ and $\epsilon \in [10^{-5},10^{-4}]$. When $\lambda_s \geq 100$, $\mu \geq 100$ or  $\epsilon \geq 10^{-3}$, the regularization is unreasonably strong. When $\lambda_s \leq 0.1$, $\mu \leq 0.1$ or $\epsilon <= 10^{-6}$, the regularization is unreasonably weak.
    \item The algorithm is not sensitive to $\sigma$, any $\sigma \leq \epsilon$ works well. 
    \item $p=\infty$ makes the size of perturbation constraint to be the same regardless of the number of dimensions.  For $p=2$, adversarial perturbation is sensitive to the number of dimensions (A higher dimension usually requires a larger perturbation), especially for sentences with different length. As a result, we need to make less tuning effort for $p=\infty$. For other values of $p$, the associated projections are computationally inefficient.
    \item We observed a minor improvement by using a larger $S$ or a larger $T_{\tilde x}$. The minor improvement comes with an increased cost of computation. When $S=T_{\tilde x}=1$, SMART requires 3 more forward passes and 3 more backward passes per iteration, compared with direct fine-tuning. In practice, it takes about 3 times the original training time. In terms of memory usage, it approximately doubles the GPU memory usage.

\item We set $\beta = 0.99$ for the first $10\%$ of the updates ($t<=0.1T$) and $\beta = 0.999$ for the rest of the updates  ($t>0.1T$) following \citep{tarvainen2017mean}, which works well in practice. 
\end{itemize}

\end{document}